\documentclass[a4paper]{article}

\usepackage{INTERSPEECH2018}
\usepackage{amssymb}
\usepackage{graphicx}
\usepackage{amsmath,epsfig}

\title{Semi-supervised and active-learning scenarios: Efficient acoustic model refinement for a low resource Indian language}
\name{Maharajan Chellapriyadharshini, Anoop Toffy, Srinivasa Raghavan K. M, V. Ramasubramanian}
\address{
	International Institute of Information Technology – Bangalore (IIIT-B), Bangalore, India}
\email{\{chellapriyadharshini.m, anoop.toffy, srinivasaraghavan.km\}@iiitb.org, v.ramasubramanian@iiitb.ac.in}

\begin{document}

\maketitle
\begin{abstract}
	We address the problem of efficient acoustic-model refinement (continuous retraining) using semi-supervised and active learning for a low resource Indian language, wherein the low resource constraints are having i) a small labeled corpus from which to train a baseline `seed' acoustic model, and ii) a large training corpus without orthographic labeling or from which to perform a data selection for manual labeling at low costs. The proposed semi-supervised learning decodes the unlabeled large training corpus using the seed model and through various protocols, selects the decoded utterances with high reliability using confidence levels (that correlate to the WER of the decoded utterances) and iterative bootstrapping. The proposed active learning protocol uses confidence level based metric to select the decoded utterances from the large unlabeled corpus for further labeling. The semi-supervised learning protocols can offer a WER reduction, from a poorly trained seed model, by as much as 50\% of the best WER-reduction realizable from the seed model's WER, if the large corpus were labeled and used for acoustic-model training. The active learning protocols allow that only 60\% of the entire training corpus be manually labeled, to reach the same performance as the entire data.
\end{abstract}


\vspace{-5pt}
\section{Introduction}

Present day speech recognition has benefited from deep learning techniques, which call for very large training corpus for training the acoustic models. A majority of languages for which speech recognition technologies are developed and deployed enjoy easily available large speech and language resources and thereby permit training of deep learning based acoustic models. While this is so, there are an equal number of diverse languages which qualify to be called low-resource languages according to several criteria. Such criteria include limited availability of digital spoken language corpus, lack of script level representations (needed for acoustic model training via labeling), limited means of labeling the speech corpus (orthographic transcrption), limited access to linguistic knowledge, expertise or resources by which to acquire lexical representations, annotations etc.


Within this spectrum of low-resource criteria, we specifically address the scenario where there is availability of adequate speech corpus, but having the data annotated (orthographic labeling) is expensive or not possible. Interestingly, such a low-resource setting has a parallel to high-resource settings such as voice-search applications (for high-resource languages), where it is required to have continuous re-training of deep-learning based acoustic models from user-data available in a continuing basis, but which are expensive to be labeled, due to the high throughput of the incoming data, which makes it difficult for a manual process to label such large volume data in a continuous manner. Such applications, requiring the incoming data to be labeled, call for techniques similar to that needed for low-resource settings where possible large speech corpus has to be labeled even for the first level acoustic model training.

Here, we address the scenario of optimally utilizing a large speech corpus without the associated orthographic labeling, by means of semi-supervised learning and active learning protocols, by which the corpus can be labeled and used for acoustic model training. The broad frameworks of semi-supervised learning and active learning has a long history in both machine learning in general 
\cite{Zhu}, \cite{Settles1}, \cite{Settles2}, \cite{Settles3}, \cite{Olsson}
 and particularly in speech recognition \cite{LiDeng}, \cite{Zhang1}, \cite{Zhang2}, \cite{Drugman}, \cite{Emre}, 
 \cite{asr1, asr2, asr3, asr4}.
 With respect to speech recognition, the early variants of semi-supervised learning were in the form of lightly-supervised acoustic model training
 \cite{Light1, Light2, Light3, Light4, Light5, Light6, Light7, Light8}, and more recently has attracted renewed attention with the requirements arising from voice-search type of applications such as referred above \cite{Google1, Google2} and low resource setting (as is the focus here)
 \cite{low-res1, low-res2, low-res3, low-res4, low-res5}. Active learning has its origins in machine learning theory \cite{Settles3}, further adapted to speech recognition in specific forms such as uncertainty sampling using confidence levels, entropy and sub-modular function based data selection 
 \cite{al1, al2, al3, al4}.
 Semi-supervised methods focus on how the unlabeled corpus can be decoded, with associated decoding errors (given the need to start with a poorly trained model with which to decode the larger corpus) and arrive at means of utilizing such erroneous decoding effectively for model improvement. Active learning methods focus on being able to select data from the decoded large corpus, in such a way that the selected data is most informative in the sense that this data has information complementary to the current acoustic model, and which therefore, when used for model retraining, offers the best model refinement comparable to the entire data.

In this paper, in the case of semi-supervised learning, we start with a seed model trained on a very low seed training data and use it to decode the large unlabeled data and propose an iterative bootstrapping protocol for using such decoded labels to efficiently retrain the acoustic models, thus completely circumventing the need for manually labeling the large unlabeled corpus. In the case of active learning, we likewise use the seed model to decode the unlabeled corpus, but perform a `data selection' by confidence level criteria, wherein the selected data can be manually labeled, and used for acoustic model training; here the focus is on showing that such a data selection can yield a smaller proportion of the entire data to be manually labeled, but offer the same performance as the entire data would, thereby resulting in a large saving in the manual effort and cost needed to reach a specific performance on a held out test corpus. 

\vspace{-5pt}
\section{Data corpus and experimental settings}

We have used a data set of an Indian language `Tamil' \footnote{Tamil language read speech data provided by SpeechOcean and Microsoft for the “Low Resource Speech Recognition Challenge for Indian Languages” in Interspeech 2018; with the lexicon IIT-Madras Common Label Set Lexicon for Tamil (57745 words + SIL + $<$UNK$>$)}. We use 15.6 hours of this data set, which is divided into three parts, as above, namely, $D_{seed}$ the `labeled' seed data set from which the seed acoustic model $AM_{seed}$ is trained, $D_U$ the unlabeled larger data set which is to be used by semi-supervised learning, and $T$ the held out test data on which to perform the test decoding to derive the WER to characterize the efficacy of the acoustic model retrained by the semi-supervised learning on $D_U$. The 15.6 hours of total data is divided into $D_{seed}$: $D_U$: $T$ in a 25:65:10 split, i.e., $D_{seed}$ is 3.74 hours, $D_U$ is 9.8 hours and $T$ is 1.52 hours.

In this work, we have used DNN-HMM framework trained using Kaldi \cite{Kaldi} and with word level trigram language models. We use MFCC feature vectors (13 dimensions spliced with +/- 4 frames to get 40 dimensional feature (MFCC + LDA + MLLT + fMLLR), spliced with +/- 5 frames to get 440 dimension feature). We use RBM pretraining (CD-1) on a train 90\% and cross-validation 10\% split and train 4 RBMs with hidden layer dimension of 1024 using 10 epochs. DNNs are trained into triphone states with 10-12 epochs of cross-entropy training and mini-batch stochastic gradient descent, with input dimension 1024, hidden layer dimension 2016 and softmax layer output dimension of 2016. We also used sequence discriminative training using state-level minimum Bayes risk (sMBR) criterion using 6 epochs of stochastic gradient descent.

\vspace{-5pt}
\section{Semi-supervised learning}

The semi-supervised learning scenario essentially involves starting with a seed acoustic model trained from a small seed data with labeling (assumed available in a low resource setting), and be able to use a significantly larger data set without labeling (as is typical in a low resource setting) and establish means of using the unlabeled data in the larger data set to refine (retrain) the seed acoustic model in such a way that the resulting refined acoustic model performs on a held out test data with performances close to what would be obtained if the acoustic model were trained with the larger data labeled with ground-truth, i.e., on all of the available data, including the seed data and the larger data, with the larger data now being labeled - for purposes of establishing the best performance realizable from the entire available data as in a high-resource setting. This belongs to the class of `self-training' approaches.

\begin{figure}[t]
	\centering
	\includegraphics[width=3.2in]{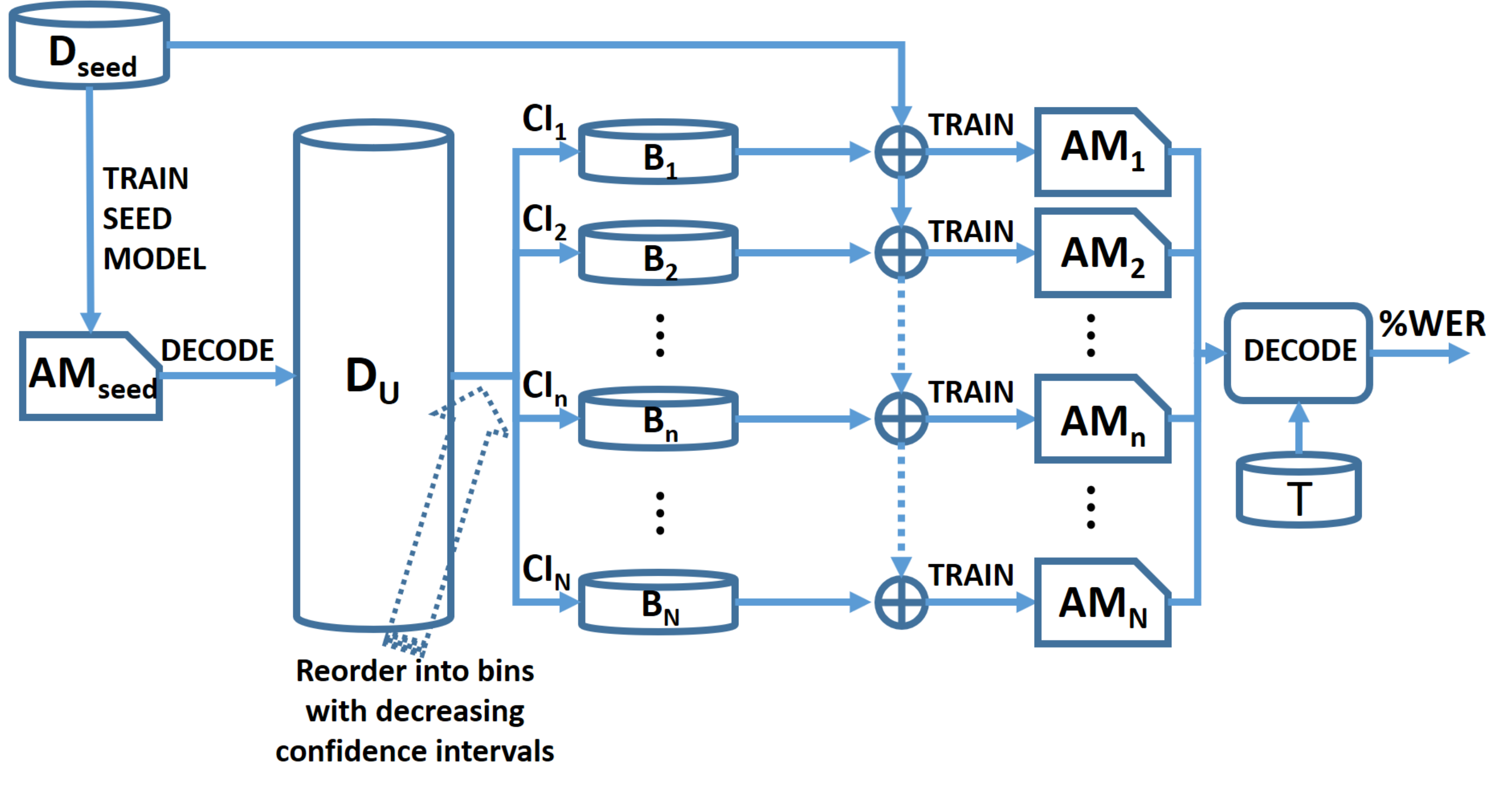}
	\vspace{-17pt}
	\caption{Framework for semi-supervised learning}
	\label{fig:BD1}
\end{figure}

We propose a broad framework of semi-supervised learning as illustrated in Fig. \ref{fig:BD1}, within which we identify two specific scenarios, a non-iterative procedure and an iterative procedure. 

\vspace{-5pt}
\subsection{Non-iterative procedure}

\begin{figure}[t]
	\centering
	\includegraphics[width=2.8in]{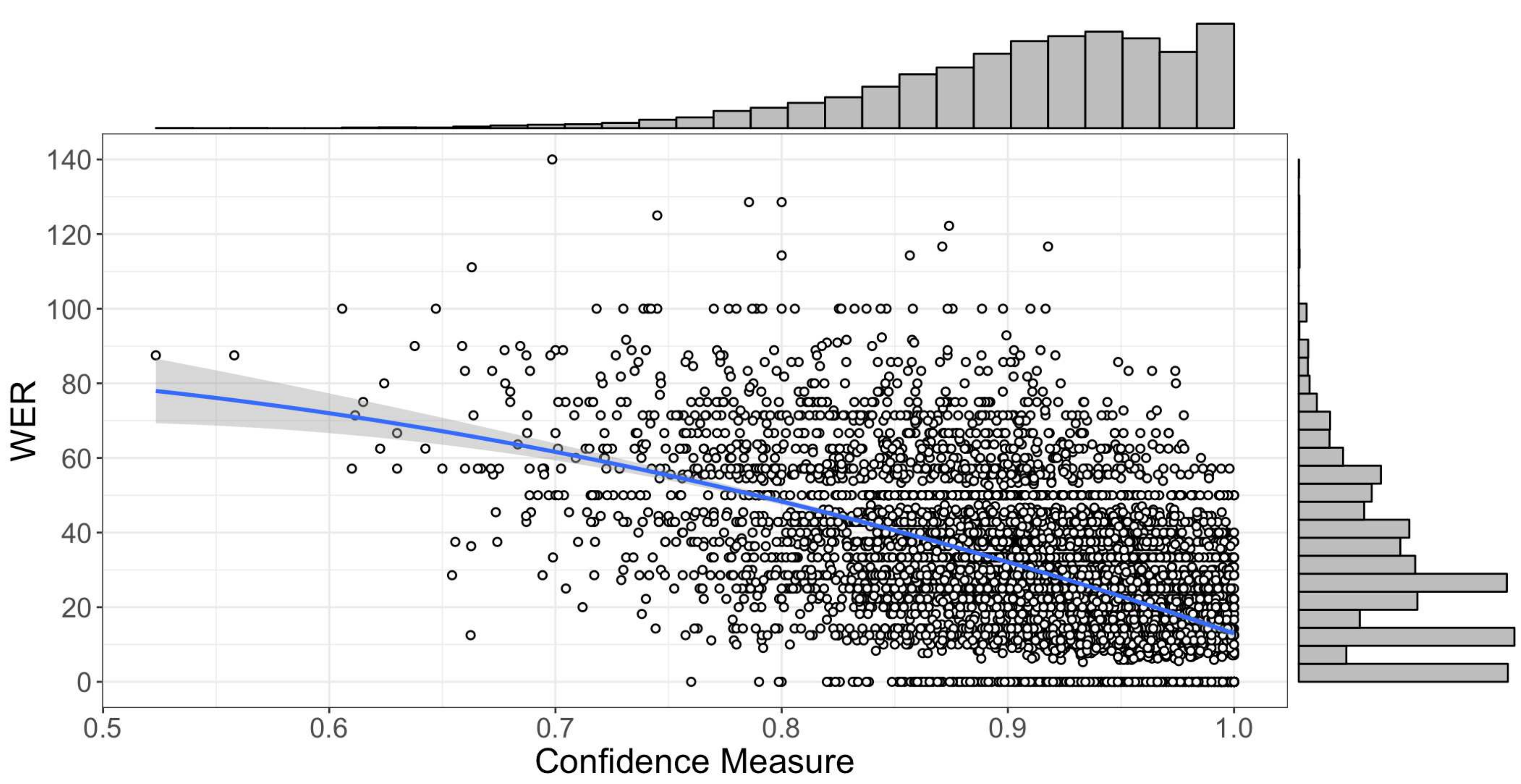}
	\vspace{-4pt}
	\caption{Scatter plot of utterance level confidence level and WER and associated confidence level and WER histograms}
	\label{fig:Scatter}
\vspace{-20pt}
\end{figure}

The semi-supervised learning scenario we consider first is called the `non-iterative' procedure as in Fig. \ref{fig:BD1}. Here, the seed acoustic model $AM_{seed}$ is used to decode $D_U$ to derive word label sequences, with an inherent WER distributed across the utterances of the data set. The WER distribution is shown in Fig. \ref{fig:Scatter}, as the vertical histogram on the WER axis ($y$-axis); it is clear that the seed acoustic model has a large spread of WER. Ideally, utterances with lower WER can be treated as close to ground truth labels and used for retraining $AM_{seed}$, thereby making available more data from $D_U$ to improve $AM_{seed}$. Since the WER is not available (as the ground truth of $D_U$ is realistically not available), we need other metrics by which we can measure the accuracy of the decoding of the utterances in $D_U$. One of the readily available measure is the confidence level of a decoded utterance, derived from the posterior of each word segment with respect to the word-level label it is aligned to in the decoding. Fig. \ref{fig:Scatter} shows this strong correlation between utterance level WER and confidence level for utterances from $D_U$ as decoded by $AM_{seed}$ (note the quadratic regression fit) with higher confidence levels corresponding to lower WERs. The distribution of confidence levels itself is shown as the histogram on the $x$-axis; typically, this is determined by how good $AM_{seed}$ is; for the $AM_{seed}$ used (from 25\% of the entire data), higher confidence levels are more likely. Note that for smaller $D_{seed}$, this distribution is likely to skew towards lower confidence levels, i.e., higher confidences will be less likely.

With the availability of the utterance level confidence level as a metric correlated to WER, we propose a `non-iterative' procedure, as in Fig. \ref{fig:BD1}. $D_U$ is split into bins $B_n, n=1, \ldots, N$ with confidence intervals $CI_n, n=1, \ldots, N$, defined by the confidence levels (0.95, 1), (0.9, 0.95), (0.85, 0.9), (0.8, 0.85), (0, 0.8) (i.e., $N=5$). The bins, in the order of decreasing confidence levels, correspond to increasing WERs and can be used to derive acoustic models $AM_n, n=1, \ldots, N$, with $AM_1$ derived from available training data, i.e., $D_{seed}+B_1$, with $B_1$ having utterances with the highest confidence levels, or lowest WERs, and likewise, $AM_n$, from $D_{seed} + B_1 + \ldots + B_n$ in an accumulative manner. Note that, with the availability of a refined $AM_1$ (from the accumulated data sets $D_{seed}+B_1$), we can generate an improved decoding of bin $B_2$ (than when derived by decoding using $AM_{seed}$ alone) and which can be used to derive $AM_2$, and likewise, $AM_{n-1}$ can be used to induce an improved decoding of bin $B_n$, whose labels are used in the training of $AM_n$.
By this overall `non-iterative' procedure, we can derive progressively refined acoustic models $AM_n, n=1, \ldots, N$, which have better decoding performance on the test data set $T$.

\begin{figure}[t]
	\vspace{-10pt}
	\centering
	\includegraphics[width=2.8in]{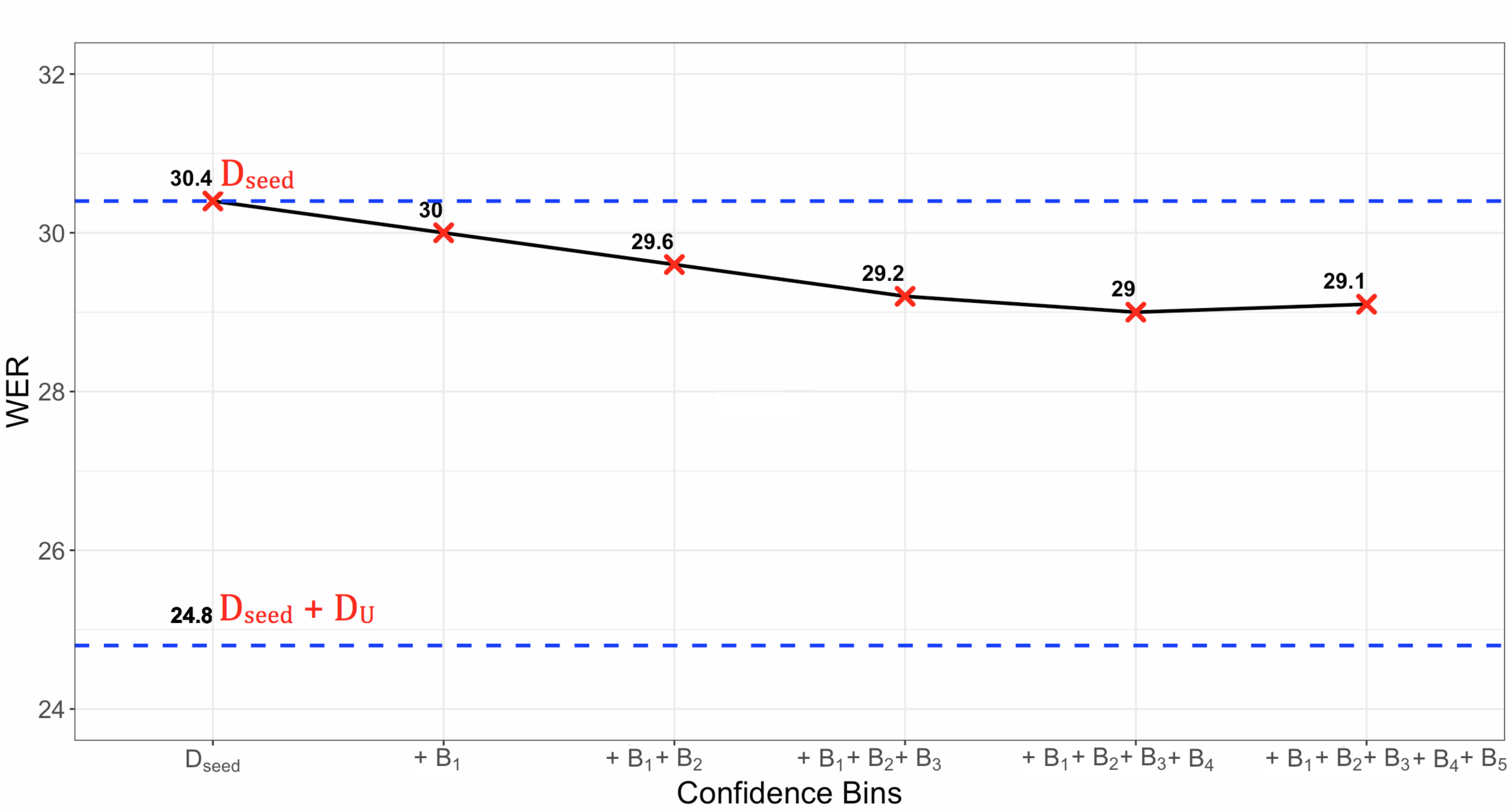}
	\vspace{-5pt}
	\caption{WER profile for the non-iterative procedure}
	\label{fig:Non-iter}
	\vspace{-20pt}
\end{figure}

Fig. \ref{fig:Non-iter} shows the actual WER profile (on $T$) for the non-iterative procedure of Fig. \ref{fig:BD1}, for the data set split $D_{seed}$: $D_U$: $T$ sets in a 25:65:10. This shows WER (on $T$) for different acoustic models: i) $AM_{seed}$ (trained with $D_{seed}$ and with a WER of 30.4\% on $T$ as shown), ii) the semi-supervised models $AM_1, \ldots, AM_n \ldots, AM_5$, and  iii) $AM_{seed}+D_U$ which is the acoustic model derived from the combined data set `$D_{seed}$ and $D_U$ with ground-truth labels' - this sets the performance limit (WER line marked $D_{seed}+D_U$ at 24.8\%) reachable by any semi-supervised protocol on $D_U$ via decoding.

\begin{figure}[t]
	\vspace{-10pt}
	\centering
	\includegraphics[width=2.5in]{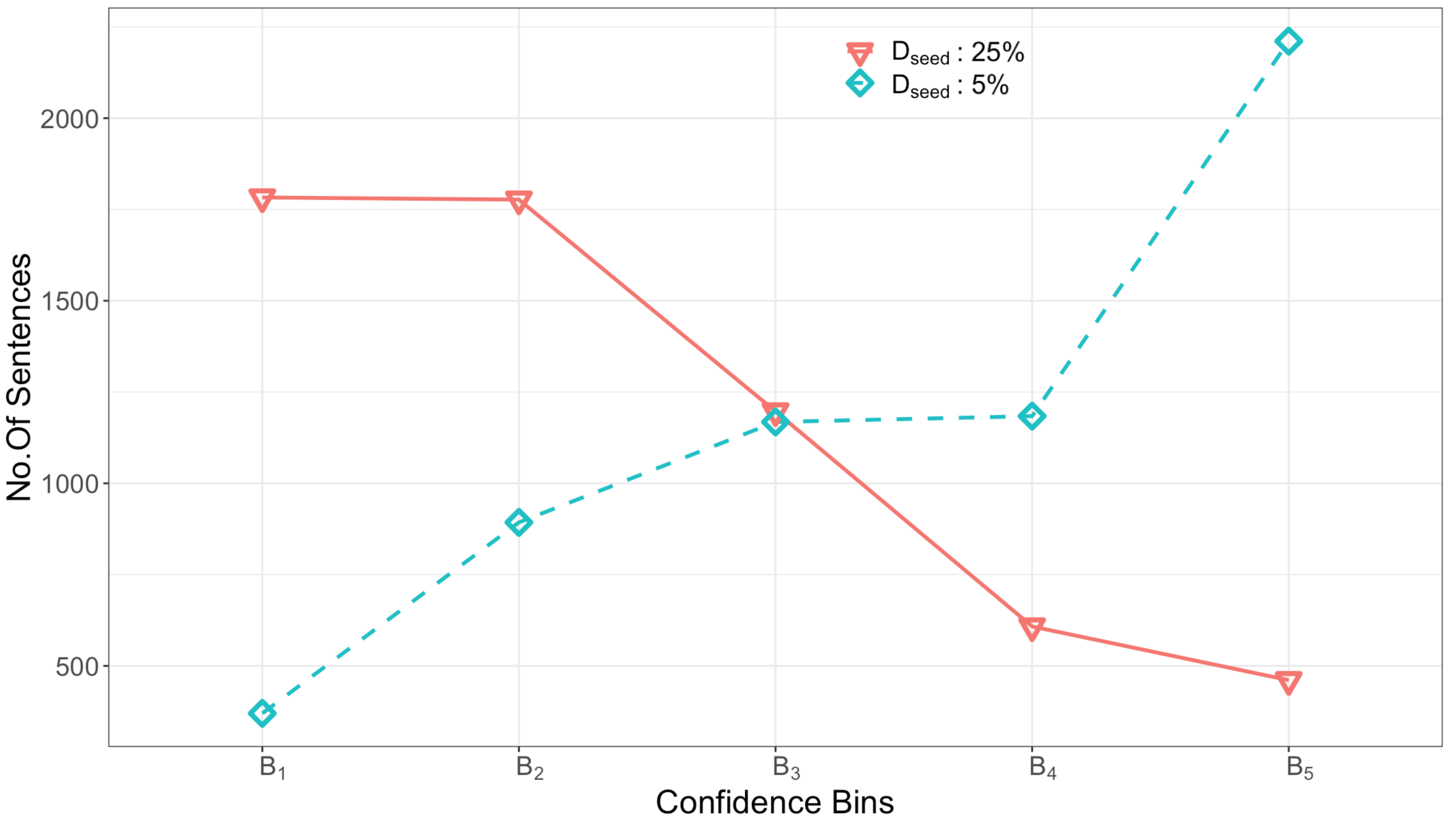}
	\vspace{-5pt}
	\caption{Distribution of confidence bins}
	\label{fig:CM-Hist-5-25}
	\vspace{-10pt}
\end{figure}

This WER profile reaches the best performance of 29\% WER up to $B_4$. This is 1.4\% lower than offered by $D_{seed}$ alone, but about 4.2\% higher than the performance limit baseline WER reachable of 24.8\%. This profile can be explained and understood in the following way: $AM_{seed}$, trained with $D_{seed}$ of 25\% split, induces a distribution of confidence levels as shown in Fig. \ref{fig:CM-Hist-5-25} (marked as $D_{seed}:25\%$. Here, the bins with a higher confidence levels are more populated, showing a good decoding for the higher confidence levels, progressively reducing for the lower confidence levels. With this bin distribution, the corresponding progressive WER in Fig. \ref{fig:Non-iter}, for $AM_{seed}$, $AM_1$, $AM_2$, \ldots, $AM_N$ shows a `convergent' behavior, i.e., the WER with addition of $B_1$ to $D_{seed}$ results in an acoustic model $AM_1$ which is `better' than $AM_{seed}$ and correspondingly lowers the WER from 30.4\%, which progressively decreases with increasing bins, until an intermediate bin (here bin $B_4$), after which the WER increases, due to the addition of the lower bin $B_5$ with lower confidence levels and hence the latter stage re-training being affected by noisy, erroneous decoded labels, causing the acoustic models to train poorly.

Note that, while this WER profile is acceptably good, a completely different (and `divergent') profile can result when $AM_{seed}$ is poorly trained, e.g. with a `very small' $D_{seed}$. This induces a distribution of confidence levels on the bins $B_n, n=1, \ldots, N$, as shown in Fig. \ref{fig:CM-Hist-5-25} (marked as $D_{seed}: 5\%$). In this distribution profile, the highest confidence level is poorly populated, with the lower confidence levels progressively more populated, resulting in bin $B_1$ having poorly decoded labels (i.e., with high WERs or erroneous labels), with successive bins also having more poorly decoded labels; as a result, the WER resulting from combining $B_1$ with $D_{seed}$ can actually show a `worsening' of WER due to highly erroneous labels - i.e., a `divergent' trend on successive additions of bins (at lesser confidence levels and increased population of utterances with higher WERs) only make this trend more acute, thereby yielding WERs higher than obtained with $AM_{seed}$, i.e., the semi-supervised learning offers no advantage and makes the performance worse. Interestingly, the result shown in Fig. \ref{fig:Non-iter} has a `non-divergent' behavior possibly attributable to the good decoding and distribution profile in Fig. \ref{fig:CM-Hist-5-25} for $D_{seed}: 25\%$).

Given this kind of `non-divergent' WER profile is realizable by the non-iterative semi-supervised learning procedure above, but with the objective of reaching WERs as close as possible to the performance limit obtainable with $D_{seed}+D_U$ i.e., the 24.8\% baseline in Fig. \ref{fig:Non-iter}, we now propose an `iterative' protocol within this broader framework.

\vspace{-5pt}
\subsection{Iterative procedure}

\begin{figure}[t]
	\centering
	\includegraphics[width=3.2in]{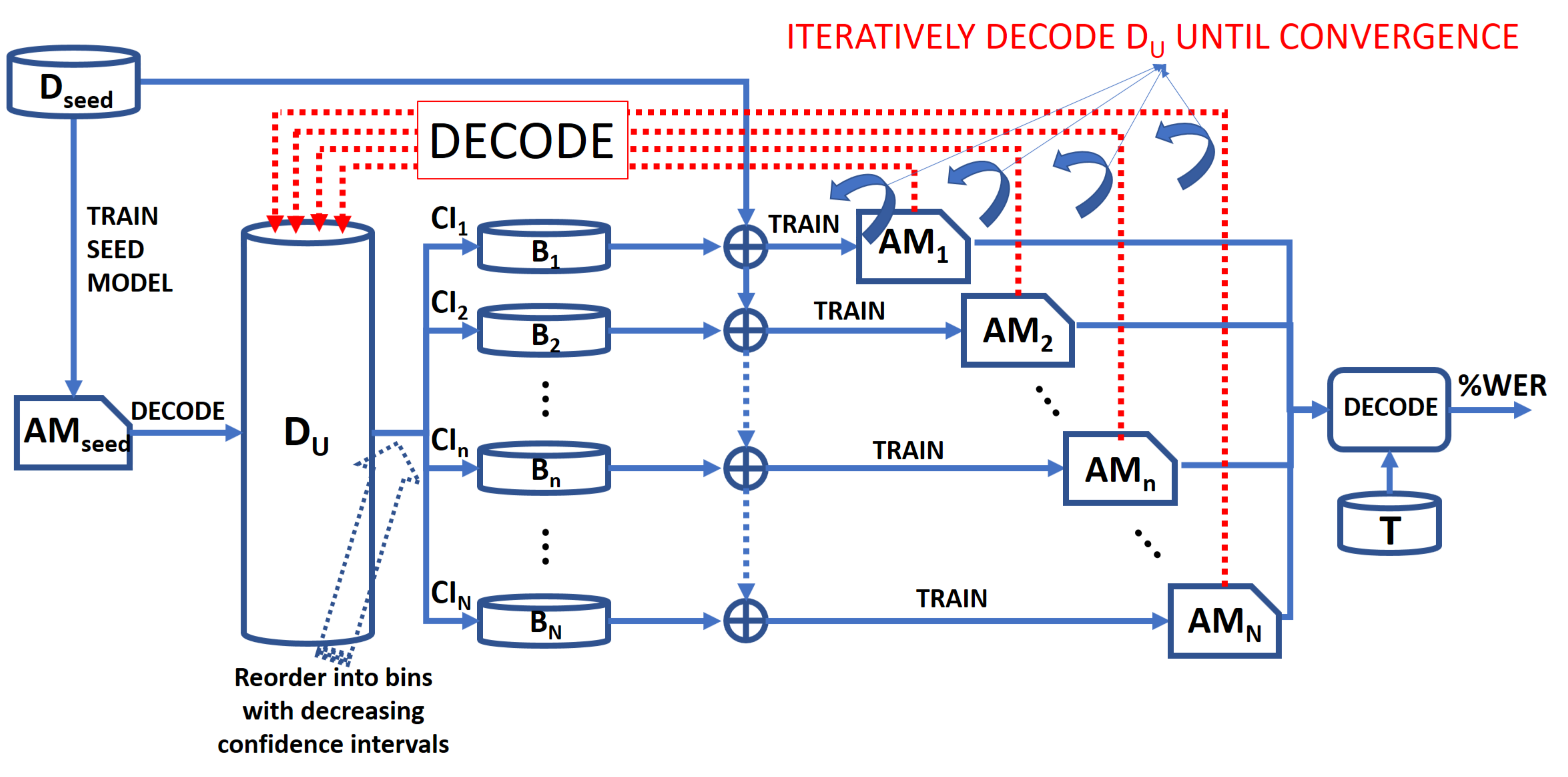}
	\vspace{-17pt}
	\caption{Semi-supervised learning: Iterative procedure}
	\label{fig:ITER-BD1}
	\vspace{-20pt}
\end{figure}

\begin{figure}[t]
	\vspace{-10pt}
	\centering
	\includegraphics[width=2.45in]{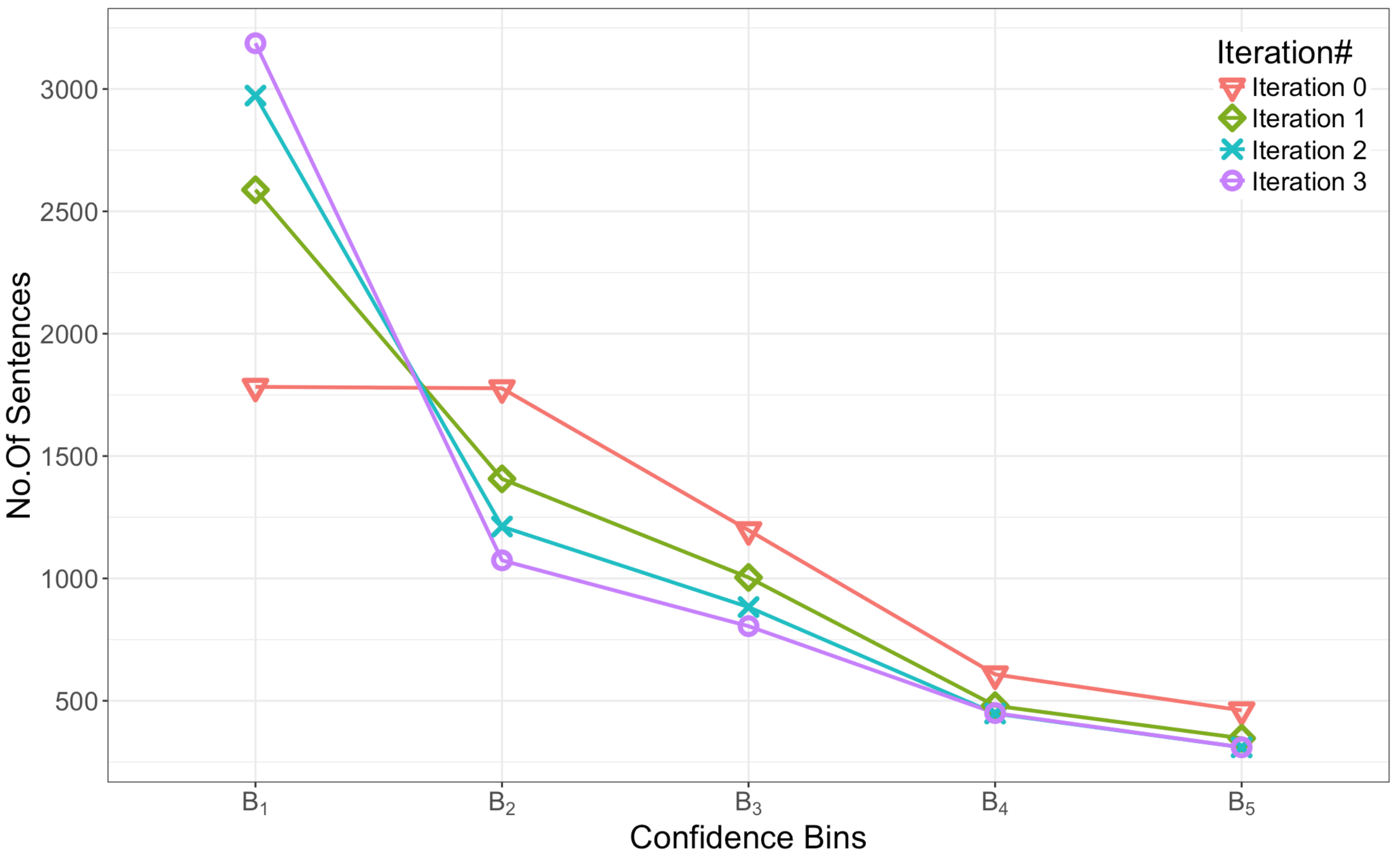}
	\vspace{-5pt}
	\caption{Iterative procedure - redistribution of utterances in bins for $AM_1$ iterations}
	\label{fig:repop}
\end{figure}

Fig. \ref{fig:ITER-BD1} shows the iterative procedure within the broad framework of Fig. \ref{fig:BD1}. Here, we primarily note that with each acoustic-model derived, $AM_n, n=1, \ldots, N$, we can decode the entire $D_U$ repeatedly to derive progressively better decoding of $D_U$ in such a way that the bins $B_n, n=1, \ldots, N$ have progressively increasing population of utterances, i.e., the utterance level confidence levels increase and the corresponding WER decrease, thereby the reuse of the iteratively refined bins results in progressively more accurate acoustic-models which in turn offer lowering of WERs on $T$. Fig. \ref{fig:ITER-BD1} shows this as iterations proceeding from each of the acoustic models $AM_n, n=1, \ldots, N$ to decode $D_U$ via (red-dashed lines via the `Decode' block). In order to visualize the effect of such an iterative decoding of $D_U$ with each $AM_n$, we show in Fig. \ref{fig:repop} for $AM_1$ iteration, the distribution of utterances in bins $B_n, n=1, \ldots, 5$ with iterations 0 to 3, with iteration-0 corresponding to the distribution obtained with decoding by $AM_{seed}$ and further iterations by decoding with $AM_1$ iteratively retrained using $D_{seed}+B_1$, with $B_1$ progressively being repopulated as shown. The progressive movement of utterances from the lower confidence bins to the higher confidence bins can be noted from Fig. \ref{fig:repop}. This in turn creates a positive `boot-strap' effect of progressive refinement of $AM_1$ from progressively more number of better decoded utterances in $B_1$. The same effect continues in the iterations corresponding to $AM_2, \ldots, AM_5$.

\begin{figure}[t]
	\vspace{-7pt}
	\centering
	\includegraphics[width=2.8in]{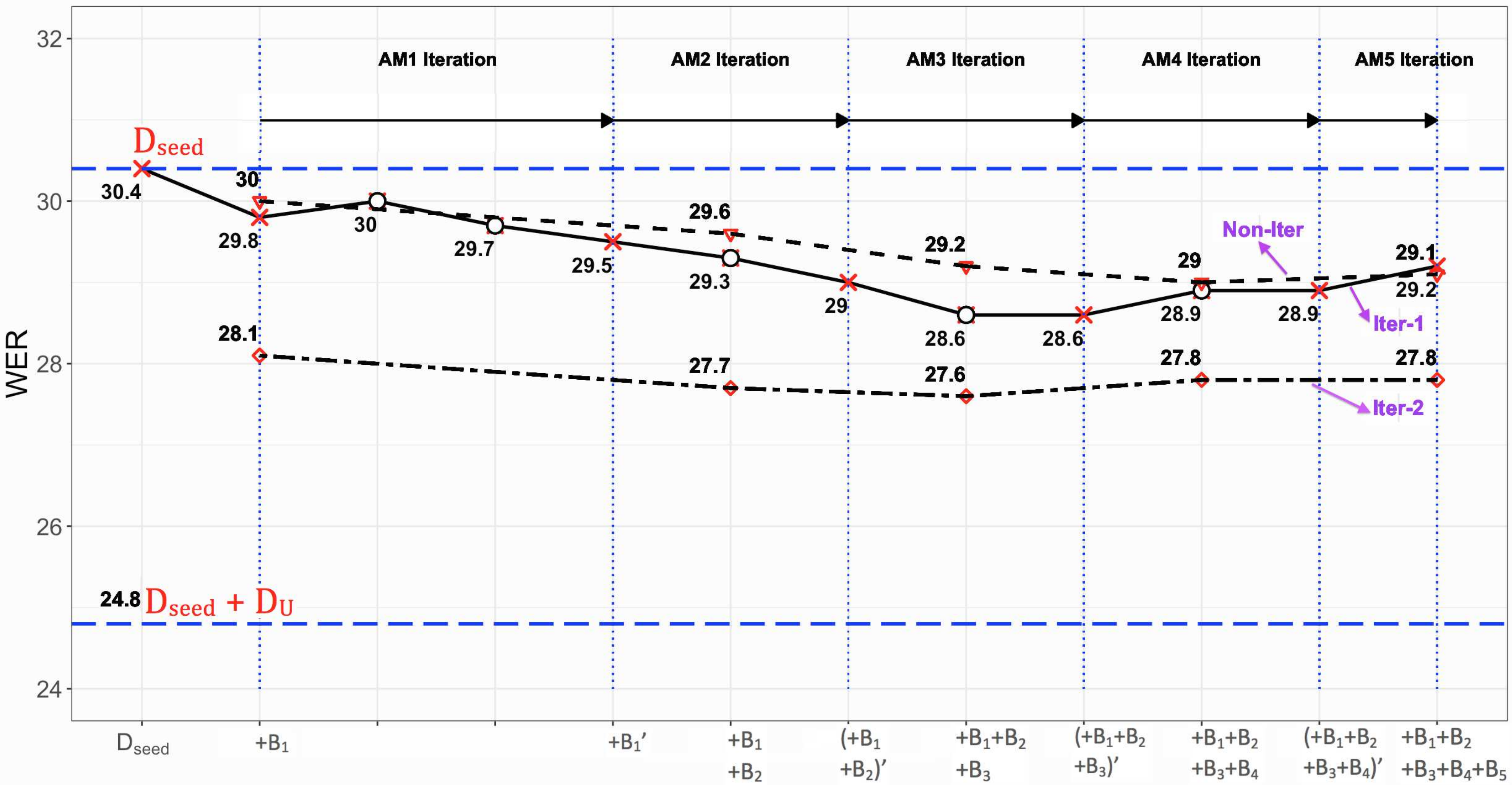}
	\vspace{-5pt}
	\caption{Iterative procedure - WER profiles}
	\label{fig:Iter-results}
	\vspace{-20pt}
\end{figure}

Fig. \ref{fig:Iter-results} shows the WER profile (marked Iter-1) for the iterative procedure described above. Each of the localized iterations for each $AM_n$ are marked. The iterative procedure yields a lower WER profile than the non-iterative procedure (Fig. \ref{fig:Non-iter}, reproduced here for comparison) with the lowest WER of 28.6\%.

Now, considering that $AM_3$ resulting from the above iterative procedure offer the lowest WER, it is possible to carry out a `global' iteration of continuing with the entire `iterative' protocol, by replacing $AM_{seed}$ with $AM_3$ to decode $D_U$ afresh, and perform another iteration as described above. This then yields a second WER profile in Fig. \ref{fig:Iter-results} (marked Iter-2), which can be seen to offer a further significant decrease in the WERs and reaching a lowest WER of 27.6\% also at $B_3$. While such `global' reiterations can be continued until the successive WER profiles converge and offer no improvements, we note that the `redistribution' of bins saturates which in turn leads to saturation of the global iterative improvememts.

In summary, we note the following from Fig. \ref{fig:Iter-results}. In the proposed semi-supervised learning framework on the Indian language Tamil, with a data-set of 15.6 hours, 3.74 hours of the data was used as labeled seed data from which the seed acoustic model $AM_{seed}$ is trained. The use of $AM_{seed}$ offers a WER of 30.4\% on the held out test data $T$. The use of acoustic model derived from $D_{seed} + D_U$, with $D_U$'s ground-truth labels, yields the performance limit of WER 24.8\%, being about 5.6\% lower than with the seed model alone. The best semi-supervised performance (with the iterative procedure, applied twice globally) offers a WER of 27.6\%, which is 2.8\% lower than with the seed model alone, and 2.8\% to the performance limit, i.e., half-way down. This is a 50\% decrease in WER of the total `WER reduction' achievable in limit, were the entire $D_U$ with ground-truth labels used for training an acoutic model. By this, the use of the proposed semi-supervised learning procedure, when applied to a large unlabeled corpus, can reduce the WER from a poorly trained seed model, 50\% closer to the WER realizable if the large corpus were labeled and used for acoustic-model training. In terms of overall absolute WER, the proposed semi-supervised learning offers a 2.8\% (absolute) reduction in WER, which is a 9\% relative reduction from the seed model's performance, without requiring labeling of the larger data set.

\vspace{-5pt}
\section{Active learning}
\vspace{-2pt}

In a complementary approach to semi-supervised learning, active learning chooses utterances from the decoded $D_U$ (large unlabeled data) that have the potential to be most informative to enhance the acoustic model (initially trained from the small seed data set $D_{seed}$). The utterances which have higher WER (indicated by lower confidence levels in the scatter plot of Fig. \ref{fig:Scatter}) are considered more informative in the sense of having phonetic content which are decoded poorly by their corresponding current acoustic models, and which therefore, when included into the retraining data - with ground truth labels - has the most potential to improve these poor acoustic models. Note that such an active learning scenario essentially performs `data selection' wherein some criterion is used to select data that is most informative, for further human labeling. The objective is to minimize the amount of data thus selected to reach a specific performance (as would be reached by the entire unlabeled data if it were to be labeled manually), in turn minimizing the manual labeling effort and costs. This is relevant in a low-resource setting, where it is desirable to minimize such efforts of manual labeling to reach a desired performance from a larger unlabeled corpus.

Towards this, we use the confidence level as the metric by which to choose the most informative utterances, in the same framework as in Fig. \ref{fig:BD1}, where the bins $B_n, n=1, \ldots, N$ are ordered in increasing confidence levels. By this, if we were to start with a poor seed model, the confidence level distribution will be skewed to have large population in lower confidence levels (since the model is poor, it induces a poor decoding of $D_U$); this in turn allows more complementary data to be manually labeled, allowing the corresponding acoustic models $AM_n, n=1, \ldots, N$ to be trained more effectively.

\begin{figure}[t]
	\vspace{-10pt}
	\centering
	\includegraphics[width=2.6in]{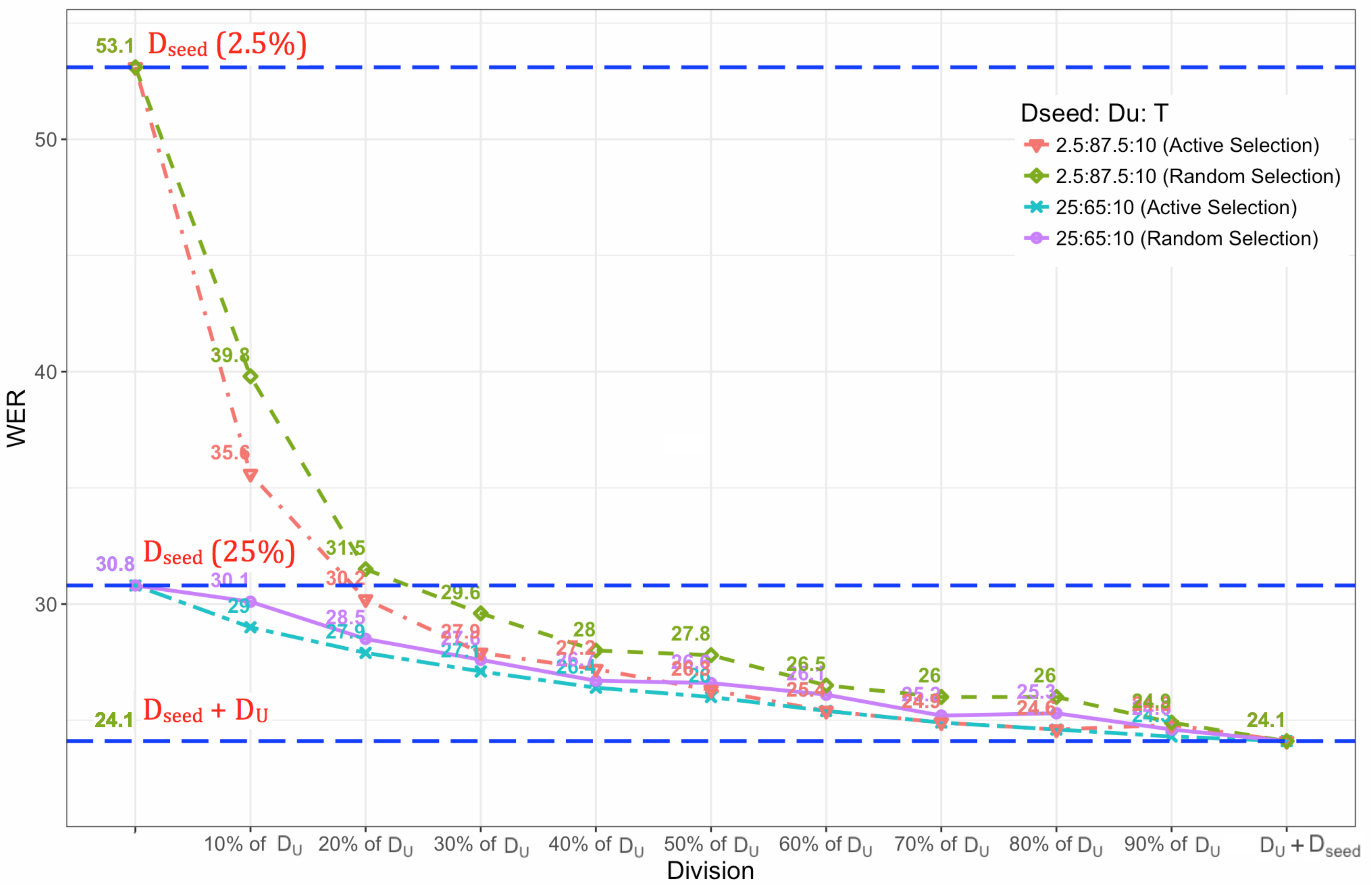}
	\vspace{-5pt}
	\caption{Active learning - WER profiles}
	\label{fig:al-results}
	\vspace{-22pt}
\end{figure}

The performance profile (WER profile) of this active learning protocol is shown in Fig. \ref{fig:al-results} for data splits $D_{seed}$: $D_U$: $T$ in a 25:65:10 and $D_{seed}$: $D_U$: $T$ in a 2.5:87.5:10. It can be seen that the 2.5:87.5:10 split has a poor $D_{seed}$ WER (53.1\%) which drops steeply with addition of the bins (in the order of increasing confidence levels), and reaches a performance of 25.4\% which is within 1\% of the performance limit of 24.1\% (with $D_{seed}$ + $D_U$) at 60\% of the full data; this is about 1\% lower than the same percent of data selected randomly. Likewise, the 25:65:10 split shows a similar trend, but given that the $D_{seed}$ is larger, the initial $D_{seed}$ WER is lower (30.8\%) and drops more gradually with increasing bins, also reaching within 1\% of the performance limit of 24.1\% at 60\% of the full data; this is less than 1\% lower than a random selection for the same amount of data selected. In summary, active learning as proposed here using confidence level based data selection does offer a performance comparable to the full data, at about 60\% of the full data, making it a useful protocol to follow in a low resource setting.

\vspace{-5pt}
\section{Conclusions}
\vspace{-2pt}

We have addressed the problem of acoustic model training in a low resource setting, where only a small seed data is assumed to be available, and have
 proposed semi-supervised learning and active learning protocols for refining the seed acoustic model from a larger, but unlabeled, training corpus. The proposed semi-supervised learning offers WER reductions by as much as 50\% of the best WER-reduction realizable from the seed model's WER, if the large corpus were labeled and used for acoustic-model training. The active learning protocols allow reduction of manual labeling to only 60\% of the entire training corpus to reach the same performance as the entire data.



\end{document}